\documentclass[10pt,twocolumn,letterpaper]{article}

\usepackage{cvpr}              

\usepackage{multirow}
\usepackage{multicol}

\newcommand{\mybold}[1]{\boldsymbol{#1}}

\definecolor{cvprblue}{rgb}{0.21,0.49,0.74}
\usepackage[pagebackref,breaklinks,colorlinks,allcolors=cvprblue]{hyperref}

\def\paperID{18142} 
\def\confName{CVPR}
\def\confYear{2025}

\title{PatchDPO: Patch-level DPO for Finetuning-free Personalized Image Generation}

\author{
{Qihan Huang}\textsuperscript{1}, {Weilong Dai}\textsuperscript{2},
{Jinlong Liu}\textsuperscript{2}, {Wanggui He}\textsuperscript{2}, {Hao Jiang}\textsuperscript{2}, {Mingli Song}\textsuperscript{1}, {Jie Song}\textsuperscript{1,$\dagger$}\\
    \textsuperscript{1} Zhejiang University,
    \textsuperscript{2} Alibaba Group\\
    {\tt\small \{qh.huang,sjie,brooksong\}@zju.edu.cn,} \\
    {\tt\small \{chenlong0104.chen,aoshu.jh\}@alibaba-inc.com,} \\
    {\tt\small LJLwykqh@126.com,wanggui.hwg@taobao.com}
}

\begin{document}
\maketitle

\footnotetext[1]{$\dagger$ Corresponding author.}

\begin{abstract}
Finetuning-free personalized image generation can synthesize customized images without test-time finetuning, attracting wide research interest owing to its high efficiency.
Current finetuning-free methods simply adopt a single training stage with a simple image reconstruction task, and they typically generate low-quality images inconsistent with the reference images during test-time.
To mitigate this problem, inspired by the recent DPO~(\ie, direct preference optimization) technique, this work proposes an additional training stage to improve the pre-trained personalized generation models.
However, traditional DPO only determines the overall superiority or inferiority of two samples, which is not suitable for personalized image generation because the generated images are commonly inconsistent with the reference images only in some local image patches.
To tackle this problem, this work proposes PatchDPO that estimates the quality of image patches within each generated image and accordingly trains the model.
To this end, PatchDPO first leverages the pre-trained vision model with a proposed self-supervised training method to estimate the patch quality.
Next, PatchDPO adopts a weighted training approach to train the model with the estimated patch quality, which rewards the image patches with high quality while penalizing the image patches with low quality.
Experiment results demonstrate that PatchDPO significantly improves the performance of multiple pre-trained personalized generation models, and achieves state-of-the-art performance on both single-object and multi-object personalized image generation.
Our code is available at \textit{~\url{https://github.com/hqhQAQ/PatchDPO}}.

\end{abstract}

\section{Introduction \label{sec:intro}}

Personalized image generation methods have garnered significant research interest, which generate images based on reference images that define specific details of the desired output.
The methodology in this domain is progressively evolving from a \textit{finetuning-based} approach~(\eg, DreamBooth~\cite{ruiz2023dreambooth}, Custom Diffusion~\cite{kumari2023custom_diffusion}) towards a \textit{finetuning-free} approach~(\textit{e.g.}, IP-Adapter~\cite{ye2023ip}, Subject-Diffusion~\cite{ma2024subject_diffusion}, JeDi~\cite{zeng2024jedi}), as finetuning-free methods eliminate the need for finetuning during test-time, significantly reducing usage costs.

\begin{figure}[t]
\centering
    \includegraphics[width=\linewidth]{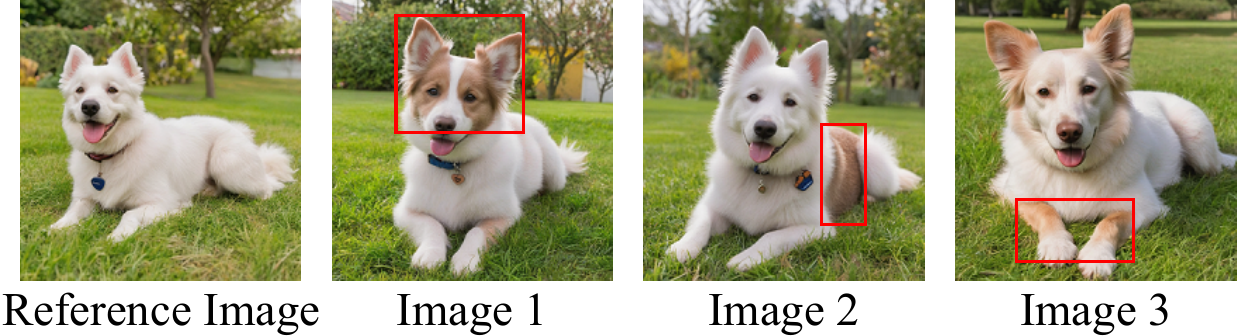}
\caption{
The generated images~(Images 1 \& 2 \& 3) are commonly inconsistent with the reference image only in some local image patches~(marked in the red boxes).
}
\vspace{-1em}
\label{fig:intro_example}
\end{figure}

Current finetuning-free methods typically employ only a single training stage on a large-scale image dataset.
During this stage, the model is trained with a simple image reconstruction task~(\ie, taking each image as reference image to reconstruct itself).
When generating images in different scenes from the reference images in test-time, existing models often generate images of lower quality~(\ie, inconsistent with the reference images in local details).

Inspired by the recent DPO technique~(\ie, direct preference optimization~\cite{rafael2023dpo}) that leverages human feedback to optimize the pre-trained model, in this work we strive to devise an \textbf{additional} training stage for improving the pre-trained personalized generation models.
Specifically, the DPO technique assigns human preference to each sample, and trains the model to generate outputs that align more closely with human preferences.
However, traditional DPO only judges the overall superiority or inferiority of two samples, which is not suitable for personalized image generation because generated images are usually inconsistent with the reference images only in some local areas, leading to inaccuracies when comparing the overall quality of two images. 
For example, as shown in \autoref{fig:intro_example}, the generated images~(1 \& 2 \& 3) are inconsistent with the reference image only in the head \& back \& leg, respectively.
In this case, traditional DPO roughly categorizes these images into superior and inferior samples, which would lead to the model's predictions \textbf{incorrectly} approaching the low-quality parts in the superior sample while moving away from the high-quality parts in the inferior sample.

To tackle this problem, this work proposes PatchDPO, which estimates the quality~(preference level) of each patch in the generated image and accordingly optimizes the model.
PatchDPO can provide feedback for the model in a more refined way, enabling the model to retain high-quality patches within images while moving away from low-quality patches.
To this end, PatchDPO can be divided into three main stages: (1) Data construction; (2) Patch quality estimation; (3) Model optimization.

In the first stage~(data construction), PatchDPO requires constructing a training dataset that includes multiple pairs of reference image and generated image.
Our preliminary experiments~\autoref{tab:ablation_experiments} show that the complex details of objects in natural images and the confusion between objects and backgrounds hinder model training.
Therefore, this work constructs a high-quality dataset for the PatchDPO training. First, this work generates the reference images using the open-source Stable Diffusion model~\cite{robin2022ldm} with text prompts instructing the background to be cleaner.
Next, the corresponding generated images are synthesized using the pre-trained personalized generation model, with the aforementioned reference images as input.

In the second stage~(patch quality estimation), traditional DPO would require extensive labeling costs to estimate the preference level of samples.
In the case of comparing patch details between reference and generated images, thanks to the excellent pre-trained vision models, PatchDPO ingeniously utilizes the pre-trained vision model to extract patch features from reference and generated images, and evaluates the quality of patches in the generated images through patch-to-patch comparisons with those in the reference images.
Besides, due to vision models~(\eg, classification models pre-trained on ImageNet~\cite{deng2009imagenet}) typically being better at extracting overall image features instead of patch features, this work proposes a \textit{self-supervised training} method to improve patch features extraction of the original vision models.
We conduct a quantitative evaluation on the HPatches dataset~\cite{balntas2017hpatches}~(a dataset with images of the same object from different perspectives and scenes), demonstrating that our method efficiently and accurately extracts patch features for patch-to-patch comparisons.

In the third stage~(model optimization), PatchDPO utilizes the patch quality estimated from the previous stage to further train the generation model.
Specifically, PatchDPO adopts a weighted training approach, which assigns higher training weights to the image patches with higher quality, while imposing penalties on those of lower quality.
Furthermore, this work also incorporates the original reference image as the \textit{ground-truth} generated image into training.
In this manner, for the patches with lower quality in the \textit{real} generated images, we increase the training weight for their corresponding patches in the \textit{ground-truth} generated image, thus encouraging the model to correct its predictions for those low-quality patches.

We perform comprehensive experiments to validate the performance of PatchDPO.
Specifically, we apply PatchDPO to multiple pre-trained personalized generation models~(\eg, IP-Adapter, ELITE) on both single-object and multi-object personalized image generation.
Experiment results on the DreamBooth dataset~\cite{ruiz2023dreambooth} and the Concept101 dataset~\cite{kumari2023custom_diffusion} demonstrate that PatchDPO significantly improves the performance of pre-trained models.
In particular, PatchDPO achieves state-of-the-art performance on both these two tasks.

To sum up, the main contributions of this work can be summarized as follows:

$\bullet$ We construct a high-quality dataset to facilitate the PatchDPO training on personalized image generation.

$\bullet$ We propose a patch quality estimation method, which adopts the pre-trained vision models with a proposed self-supervised training approach for assessing the quality of patches in the generated images.

$\bullet$ Based on the estimated patch quality, we propose a weighted training approach for the PatchDPO training on personalized image generation.

$\bullet$ Experiment results show that PatchDPO achieves state-of-the-art performance on both single-object and multi-object personalized image generation.

\begin{figure*}[t]
\centering
    \includegraphics[width=\linewidth]{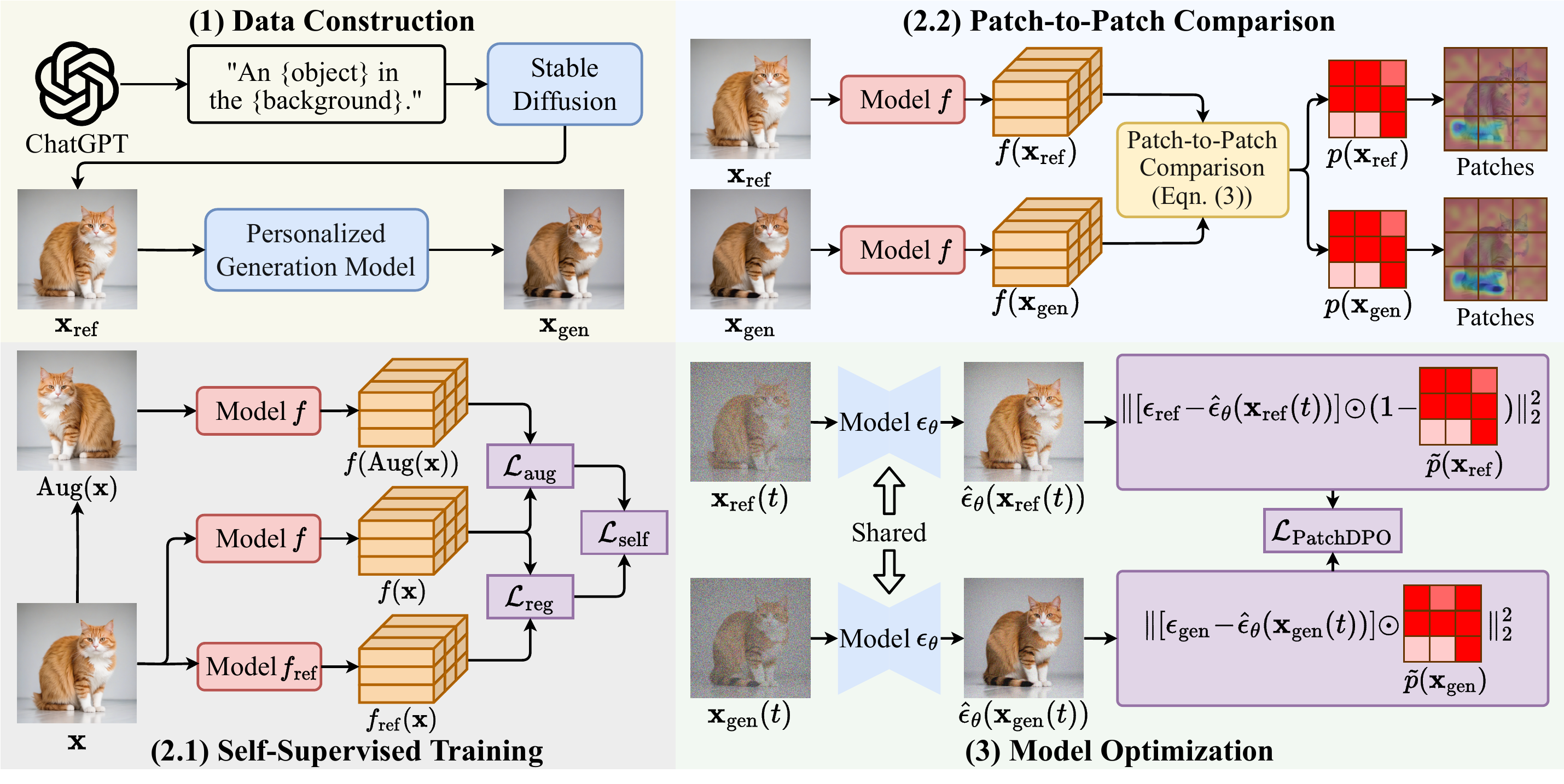}
\caption{
PatchDPO has three stages: (1) Data construction; (2) Patch quality estimation; (3) Model optimization. The stage (2) is split into (2.1) self-supervised training and (2.2) patch-to-patch comparison.
Besides, in (3), $\hat{\mybold{\epsilon}}_{\theta}({\bf x}_{\rm ref}(t))$ abbreviates $\mybold{\epsilon}_{\theta}(\mybold{x}_{\rm ref}(t), \mybold{c}_{\rm text}, \mybold{x}_{\rm ref}, t)$,
and $\hat{\mybold{\epsilon}}_{\theta}({\bf x}_{\rm gen}(t))$ abbreviates $\mybold{\epsilon}_{\theta}(\mybold{x}_{\rm gen}(t), \mybold{c}_{\rm text}, \mybold{x}_{\rm ref}, t)$.
$\tilde{p}(\mybold{x}_{\rm ref})$ and $\tilde{p}(\mybold{x}_{\rm gen})$ are upsampled from $p(\mybold{x}_{\rm ref})$ and $p(\mybold{x}_{\rm gen})$, respectively.
}
\vspace{-1em}
\label{fig:method}
\end{figure*}

\section{Related Work}

\noindent \textbf{Personalized image generation.}
Early personalized image generation methods~(\eg, DreamBooth~\cite{ruiz2023dreambooth}, Textual Inversion~\cite{gal2023textual_inversion}, Cones~\cite{liu2023cones}, Mix-of-Show~\cite{gu2023mix_of_show}) require finetuning the original diffusion model with the reference images.
Recently, finetuning-free methods~(\eg, IP-Adapter~\cite{ye2023ip}, ELITE~\cite{wei2023elite}, Subject-Diffusion~\cite{ma2024subject_diffusion}, BLIP-Diffusion~\cite{li2023blip_diffusion}, InstantBooth~\cite{shi2024instantbooth}, FastComposer~\cite{xiao2023fastcomposer}, Taming Encoder~\cite{jia2023taming}, SSR-Encoder~\cite{zhang2024ssr}, JeDi~\cite{zeng2024jedi}) emerge and attract more research interest as they require no finetuning during test-time and significantly reduce the usage cost.
However, finetuning-free methods employ only a single training stage with a simple image reconstruction task, leading to low-quality generated images inconsistent with the reference images.
PatchDPO compensates for this deficiency using an additional training stage for model optimization from the feedback.

\vspace{0.5em}
\noindent \textbf{Aligning diffusion models.}
The model alignment methods~(\eg, RLHF, DPO) first emerged in the field of large language model~(LLM).
Specifically, RLHF~(Reinforcement Learning from Human Feedback)~\cite{bai2022constitutional, menick2022teaching} trains a reward function from comparative data of model outputs to reflect human preferences, and adopts reinforcement learning to align it with the original model.
DPO~(Direct Preference Optimization)~\cite{rafael2023dpo} simplifies RLHF by aligning the original model directly on the feedback data, but matching RLHF in performance.
Recently, some methods~(\eg, DPOK~\cite{fan2023dpok}, DDPO~\cite{black2024ddpo}, Diffusion-DPO~\cite{wallace2024diffusion_dpo}, DRaFT~\cite{clark2024draft}, AlignProp~\cite{prabhudesai2023alignprop}) have integrated RLHF or DPO into diffusion models for improving image aesthetic.
However, these methods simply estimate the overall quality of the entire image, which is not suitable for personalized image generation because the generated images are usually inconsistent with the reference images only in some local image patches.
Therefore, in this work, we propose PatchDPO, an advanced model alignment method for personalized image generation by estimating patch quality instead of image quality for model training.

\section{Preliminaries \label{sec:preliminaries}}

\noindent \textbf{Diffusion model.}
Existing personalized image generation models utilize diffusion model~\cite{ho2020ddpm, rombach2022ldm} as the base model.
Diffusion model comprises two processes: a diffusion process which gradually adds noise into the original image with a Markov chain in $T$ steps, and a denoising process which predicts the noise to reconstruct the image with a deep neural network.
Detailedly, personalized image generation methods synthesize images simultaneously conditioned on the text prompt and the reference images.
Commonly, $\mybold{\epsilon}_{\theta}$ denotes the deep neural network for noise prediction, and the training loss of personalized diffusion model is calculated as below~($\| \cdot \|_{2}$ denotes the L2 norm):

\vspace{-0.5em}
\begin{equation}
\label{equa:mse_loss}
    \mathcal{L}_{\rm mse} \! = \! \mathbb{E}_{\mybold{x}_0, \mybold{\epsilon} \in \mathcal{N}(\mathbf{0}, \mathbf{I}), \mybold{c}_{\rm text}, \mybold{c}_{\rm img}} \| \mybold{\epsilon} - \mybold{\epsilon}_{\theta}(\mybold{x}(t), \mybold{c}_{\rm text}, \mybold{c}_{\rm img}, t) \|_{2}^{2},
\nonumber
\end{equation}

where $\mybold{x}_0$ denotes the original real image, $t \in [0, T]$ denotes the time step in the diffusion process,
$\mybold{x}(t) = \alpha_t \mybold{x}_0 + \sigma_t \mybold{\epsilon}$, and $\alpha_t$, $\sigma_t$ are predefined weights for step $t$ in the diffusion process.
$\mybold{c}_{\rm text}$ denotes the text condition, and $\mybold{c}_{\rm img}$ denotes the reference image.
After training, the model can generate images by gradually denoising Gaussian noise in multiple steps.

\vspace{0.5em}
\noindent \textbf{Reinforcement learning from human feedback~(RLHF).}
RLHF~\cite{bai2022constitutional, menick2022teaching} trains the model by maximizing the reward of model output, while regularizing the KL-divergence between it and the original model.
Specifically, RLHF trains a reward function $r(\mybold{c}, \mybold{x})$ that estimates the human preference on the generated sample $\mybold{x}$ conditioned on $\mybold{c}$.
Next, let $p_{\theta}$ denote the model being optimized, $p_{\rm ref}$ denote the original model, the training target of RLHF is calculated as below~(note that $\beta$ is the hyper-parameter):

\vspace{-0.5em}
\begin{equation}
\max\limits_{p_{\theta}} \mathbb{E}_{\mybold{c}, \mybold{x}} \left[ r(\mybold{c}, \mybold{x}) \right] - \beta \: \mathbb{D}_{\rm KL} \left[ p_{\theta}(\mybold{x} | \mybold{c}) || p_{\rm ref}(\mybold{x} | \mybold{c}) \right].
\end{equation}

\noindent \textbf{Direct preference optimization~(DPO).}
Direct preference optimization simplifies RLHF by training the model directly from human preferences.
Detailedly, let $\mybold{x}^{w}$ and $\mybold{x}^{l}$ denote the ``winning'' and ``losing'' samples generated from the condition $\mybold{c}$, then DPO optimizes the model by aligning its output closer to $\mybold{x}^{w}$ while distancing it from $\mybold{x}^{l}$, and the DPO loss~\cite{rafael2023dpo} $\mathcal{L}_{\rm DPO}$ is calculated as below~(note that $\sigma(\cdot)$ denotes the sigmoid function):

\vspace{-0.5em}
\begin{equation}
\label{equa:dpo_loss}
\scriptstyle
   \mathcal{L}_{\rm DPO} = - \mathbb{E}_{\mybold{c}, \mybold{x}^{w}, \mybold{x}^{l}} \left[ \log \sigma \left( \beta \log \frac{ p_{\theta}(\mybold{x}^{w} | \mybold{c}) }{ p_{\rm ref} (\mybold{x}^{w} | \mybold{c}) } - \beta \log \frac{ p_{\theta}(\mybold{x}^{l} | \mybold{c}) }{ p_{\rm ref} (\mybold{x}^{l} | \mybold{c}) } \right) \right].
\end{equation}

\section{PatchDPO}

PatchDPO consists of three stages: (1) Data construction; (2) Patch quality estimation; (3) Model optimization.

\subsection{Data Construction \label{sec:data_construction}}

PatchDPO requires constructing a training dataset comprising multiple pairs of reference image and generated image~(the generated image is synthesized by the personalized generation model being optimized).
Our preliminary experiments in \autoref{tab:ablation_experiments} demonstrate that natural images typically contain images of low-quality, which are not suitable for the task of PatchDPO.
Detailedly, these low-quality images comprise complex object details with the confused foreground and background, easily misleading the model training.
Therefore, in this work, we choose to construct a high-quality training dataset generated from the open-source generation model using three steps.

\textbf{First}, this work utilizes ChatGPT to generate the text prompt for each image.
The text prompt is in the format of ``\textit{An \{object\} in the \{background\}}'', where \{object\} and \{background\} are imagined by ChatGPT.
\textbf{Second}, this work feeds the generated text prompts into the open-source text-to-image generation model~(\eg, Stable Diffusion) to generate the reference images.
Besides, in addition to the original text prompt, the generation model is also instructed to generate simple backgrounds for the mitigation of confusion between object and background.
\textbf{Finally}, this work employs the target personalized generation model to generate images, with the aforementioned text prompts and the corresponding reference images as input.

\subsection{Patch Quality Estimation\label{sec:patch_quality_estimation}}

Traditional DPO simply estimates the overall quality of the entire generated image, which does not provide sufficiently fine and accurate feedback for personalized image generation, thus resulting in deficient performance.
To address this problem, PatchDPO estimates the quality of each patch in the generated image to acquire precise feedback for model optimization.
Besides, traditional DPO requires a large amount of annotation cost to estimate the quality of the samples.
Instead, the patch quality in personalized image generation is evaluated by comparing the patch details between reference images and generated images, which can be conducted using the pre-trained vision models.
To this end, this work proposes a \textbf{patch-to-patch comparison} method to estimate the patch quality, and proposes a \textbf{self-supervised training} method for further improvement.

\subsubsection{Patch-to-Patch Comparison}

Inspired by ProtoPNet~\cite{chen2019protopnet,huang2023evalprotopnet, huang2024concept, huang2024lgcav, xue2024protopformer} and SFD2~\cite{xue2023sfd2} that extract patch features from the deep feature maps, PatchDPO estimates the patch quality with a patch-to-patch comparison on the patch features extracted from the deep feature maps.

Specifically, let $\mybold{x}_{\rm ref}$ denote the reference image, $\mybold{x}_{\rm gen}$ denote the corresponding generated image, $f$ denote the pre-trained vision model, then $f(\mybold{x}_{\rm ref}) \in \mathbb{R}^{H \times W \times D}$ and $f(\mybold{x}_{\rm gen}) \in \mathbb{R}^{H \times W \times D}$ are the feature maps extracted by $f$~(note that $H$, $W$, $D$ are the size of the feature map).
To ensure the generalizability of PatchDPO, this work acquires the last feature maps extracted from the vision models pre-trained on ImageNet as $f(\mybold{x}_{\rm ref})$ and $f(\mybold{x}_{\rm gen})$.
Because the model does not change the spatial position of feature maps during feature extraction,
$f(\mybold{x})[h, w] \in \mathbb{R}^{D}$ represents the features of the patch $\mybold{x}[h, w]$ within the image $\mybold{x}$.
Note that $\mybold{x}[h, w]$ denotes the patch in the $h$-th row and the $w$-th column of $\mybold{x}$, as shown in the right side of \autoref{fig:method}~\textbf{(2.2)}.
Next, the quality of each patch $\mybold{x}_{\rm gen}[h, w]$ is estimated according to the existence of a patch similar to it in the reference image $\mybold{x}_{\rm ref}$.
Detailedly, the patch quality $p(\mybold{x}_{\rm gen}[h, w])$ of $\mybold{x}_{\rm gen}[h, w]$ is calculated as the maximum similarity between $f(\mybold{x}_{\rm gen})[h, w]$ with all elements in $f(\mybold{x}_{\rm ref})$:

\vspace{-0.5em}
\begin{equation}
\label{equa:patch_quality}
    p(\mybold{x}_{\rm gen}[h, w]) \! = \! \max\limits_{i,j} \frac{ f(\mybold{x}_{\rm gen})[h, w] \cdot f(\mybold{x}_{\rm ref})[i, j] }{ \| f(\mybold{x}_{\rm gen})[h, w] \| \| f(\mybold{x}_{\rm ref})[i, j] \| },
\end{equation}

where $i$, $j$ iterate over all the indexes of elements in $f(\mybold{x}_{\rm ref})$.
Therefore, higher $p(\mybold{x}_{\rm gen}[h, w])$ indicates that $\mybold{x}_{\rm gen}[h, w]$ is more consistent with the corresponding patch in the reference image $\mybold{x}_{\rm ref}$.

\vspace{0.5em}
\noindent \textbf{Verification by $S_{\rm patch}$.}
To guarantee precise patch quality estimation, this work conducts a quantitative evaluation of the extracted patch features using the HPatches dataset~\cite{balntas2017hpatches}.   
Detailedly, the HPatches dataset consists of images from 108 groups, where the images of the same group contain the same object from different perspectives and scenes.
Besides, for the same group of images, the HPatches dataset annotates the spatial correspondence between their image patches.
Based on this dataset, this work adopts a \textit{patch matching score} $S_{\rm patch}$ to evaluate the extracted patch features.
$S_{\rm patch}$ is calculated in three steps:
(1) Extract the patch features of all images in the dataset, using the pre-trained vision model $f$.
(2) For each patch in the image, predict its most similar patch~(calculated from the patch features) in other images from the same group.
(3) Calculate the matching accuracy of each patch by comparing the predicted patch with the ground-truth patch, and $S_{\rm patch}$ is finally calculated by averaging them.

As shown in \autoref{tab:hpatches_acc}~(a ViT-Base model with 12 layers is adopted here), $S_{\rm patch}$ of the patch features extracted from the last feature maps achieves only 68.4\%, which is not sufficient for patch quality estimation.

\begin{table}
\renewcommand\arraystretch{1}
\small
\centering
\setlength{\tabcolsep}{1.8mm}{
\begin{tabular}{c *5{c}}
  \toprule
\textbf{\small Model/Layer} & \textbf{\small 1} & \textbf{\small 4} & \textbf{\small 7} & \textbf{\small 10} & \textbf{\small 12} \\

\midrule
{\small ViT-Base} & 70.4 & 76.4 & 74.8 & 68.6 & 68.4 \\
{\small ViT-Base~(After Training)} & \textbf{72.7} & \textbf{82.7} & \textbf{83.7} & \textbf{77.4} & \textbf{75.8} \\

\bottomrule
\end{tabular}}
\caption{$S_{\rm patch}$~(\%) estimated on the HPatches dataset.}
\vspace{-1em}
\label{tab:hpatches_acc}

\end{table}

\subsubsection{Self-Supervised Training}


To facilitate the patch quality estimation, this work strives to improve $S_{\rm patch}$ from two aspects:
(1) Extract patch features from the shallow layers instead of the latest layer.
(2) Finetune the vision model $f$ with self-supervised training.

In the first aspect, the deep neurons in the deep neural networks have large effective receptive fields~\cite{luo2016understanding, araujo2019computing, jing2021turning, jing2023deep, huang2024resolving}, meaning that each element in the deeper feature map perceives a larger region of the image \textbf{rather than} the image patch in the corresponding location.
Therefore, this work explores extracting patch features from feature maps at different depths.
As shown in \autoref{tab:hpatches_acc}, patch features extracted from shallow feature maps have higher $S_{\rm patch}$ in general, and in particular, the patch features extracted from the 4-th layer have the highest $S_{\rm patch}$.

In the second aspect, the performance of the aforementioned patch features extraction is still limited, because the used vision models are typically trained for other vision tasks~(\eg, image classification), which focus on extracting the overall image features instead of the patch features.
Consequently, this work proposes a \textit{self-supervised training} method to finetune the pre-trained vision model $f$, towards improving the performance of patch features extraction without expensive labeling costs.
This self-supervised method augments the input image through spatial transformation~(\ie, image rotation, image flip) and then constrains the patch features at corresponding positions of the input image and the augmented image to be close.
Specifically, let ${\rm Aug}(\cdot)$ denote the augmentation operation~(\eg, ${\rm Aug}(\mybold{x})$ is the augmented image of the input image $\mybold{x}$), then the loss function $\mathcal{L}_{\rm self}$ of self-supervised training is an MSE loss with a regularization term calculated as below~($f_{\rm ref}$ denotes the original model that is frozen during training):

\vspace{-0.5em}
\begin{equation}
    \begin{cases}
        \mathcal{L}_{\rm self} = \mathcal{L}_{\rm aug} + \mathcal{L}_{\rm reg}. \\
        \mathcal{L}_{\rm aug} = \| {\rm Aug}(f(\mybold{x})) - f({\rm Aug}(\mybold{x})) \|_2^2. \\
        \mathcal{L}_{\rm reg} = \| f(\mybold{x}) - f_{\rm ref}(\mybold{x}) \|_2^2.
    \end{cases}
\end{equation}

Here, the regularization term could avoid excessive deviation of the finetuned model from the original model, stabilizing model training.
Besides, this work chooses the dataset constructed in \autoref{sec:data_construction} for this finetuning, because the finetuned vision model $f$ will be finally employed for patch quality estimation in this dataset.
After the self-supervised training, as shown in \autoref{tab:hpatches_acc}, $S_{\rm patch}$ of patch features at different layers have shown a significant improvement.
We select the patch features with the highest $S_{\rm patch}$~(83.7\%, from the 7th layer) for patch quality estimation, ensuring the performance of PatchDPO training.

\subsection{Model Optimization}

With the vision model $f$ finetuned from the previous stage, PatchDPO estimates the patch quality $p(\mybold{x}_{\rm gen}) \in \mathbb{R}^{H \times W}$ for all generated images in the training dataset.
Note that $p(\mybold{x}_{\rm gen})[h,w] = p(\mybold{x}_{\rm gen}[h,w]) \in \mathbb{R}$ is the patch quality of image patch $\mybold{x}_{\rm gen}[h, w]$.
Next, different from traditional DPO simply aligning with the superior samples while distancing from the inferior samples,
PatchDPO leverages a weighted training method that adopts a more precise approach for model optimization.
Specifically, PatchDPO trains the original personalized generation model with an image reconstruction task~(reconstructing the generated image according to the reference image), and then assigns higher training weights to the image patches with higher quality, while assigning lower training weights to the image patches with lower quality.

However, only reconstructing the generated image can lead the model to still generate low-quality patches in the generated images, instead of generating the corresponding correct patches in the reference images.
To address this problem, PatchDPO simultaneously involves a task of reconstructing the reference image using the reference image, as the ground-truth to correct the low-quality patches in the generated image.
To this end, PatchDPO estimates the patch quality $p(\mybold{x}_{\rm ref}) \in \mathbb{R}^{H \times W}$ for the reference image by comparing the features of each patch with those in the generated image, in the same manner as \autoref{equa:patch_quality}.
Like $p(\mybold{x}_{\rm gen})$, each $p(\mybold{x}_{\rm ref})[h, w]$ reflects the extent to which $\mybold{x}_{\rm ref}[h,w]$ exists in the generated image $\mybold{x}_{\rm gen}$.
Therefore, $\mybold{x}_{\rm ref}[h,w]$ with lower $p(\mybold{x}_{\rm ref})[h, w]$ indicates that the patch $\mybold{x}_{\rm ref}[h, w]$ has low generation quality in the generated image, and the training weight of this patch should be increased in the task of reconstructing the \textit{ground-truth} image~(\ie, the reference image) from the reference image.
Finally, the loss $\mathcal{L}_{\rm PatchDPO}$ of PatchDPO is calculated as below:

\vspace{-0.5em}
\begin{align}
   &\mathcal{L}_{\rm PatchDPO} \! = \! \| \! \underbrace{ \left[ \mybold{\epsilon}_{\rm gen} \! - \! \mybold{\epsilon}_{\theta}(\mybold{x}_{\rm gen}(t), \mybold{c}_{\rm text}, \mybold{x}_{\rm ref}, t) \right] }_{\rm Reconstruct \, \mybold{x}_{\rm gen} \, with \, \mybold{x}_{\rm ref}} \! \odot \, \tilde{p}(\mybold{x}_{\rm gen}) \|_{2}^{2} \nonumber \\
   &+ \| \underbrace{ \left[ \mybold{\epsilon}_{\rm ref} \! - \! \mybold{\epsilon}_{\theta}(\mybold{x}_{\rm ref}(t), \mybold{c}_{\rm text}, \mybold{x}_{\rm ref}, t) \right] }_{\rm Reconstruct \, \mybold{x}_{\rm ref} \, with \, \mybold{x}_{\rm ref}} \odot (1 - \tilde{p}(\mybold{x}_{\rm ref})) \|_{2}^{2}.  \nonumber
\end{align}

Note that $\tilde{p}(\mybold{x}_{\rm gen})$ and $\tilde{p}(\mybold{x}_{\rm ref})$ are upsampled from the original $p(\mybold{x}_{\rm gen})$ and $p(\mybold{x}_{\rm ref})$ with a normalization operation to constrain the values within $[0, 1]$, which have the same height and width as original images~($\mybold{x}_{\rm gen}$ \& $\mybold{x}_{\rm ref}$) and noise~($\mybold{\epsilon}_{\rm gen}$ \& $\mybold{\epsilon}_{\rm ref}$).
Besides, $\odot$ denotes element-wise multiplication that assigns the weights~($\tilde{p}(\mybold{x}_{\rm gen})$ \& $1 - \tilde{p}(\mybold{x}_{\rm ref})$) to the corresponding patches in the reconstruction losses.


\section{Experiments}

\begin{table}
\renewcommand\arraystretch{1}
\small
\centering
\setlength{\tabcolsep}{1.2mm}{
\begin{tabular}{c *4{c}}
  \toprule
\textbf{\small Method} & \textbf{\small DINO} & \textbf{\small CLIP-I} & \textbf{\small CLIP-T}  & \textbf{\small Avg.} \\

\midrule
{\small Real Images~\cite{ruiz2023dreambooth}} & 0.774 & 0.885 & N/A & N/A \\
{\small Textual Inversion~\cite{gal2023textual_inversion}} & 0.569 & 0.780 & 0.255 & 0.535 \\
{\small DreamBooth~(Imagen)~\cite{ruiz2023dreambooth}} & 0.696 & 0.812 & 0.306 & 0.605 \\
{\small DreamBooth~(SD)~\cite{ruiz2023dreambooth}} & 0.668 & 0.803 & 0.305 & 0.592 \\
{\small Custom Diffusion~\cite{kumari2023custom_diffusion}} & 0.643 & 0.790 & 0.305 & 0.579 \\

\midrule
{\small Re-Imagen~\cite{chen2023reimagen}} & 0.600 & 0.740 & 0.270 & 0.537 \\
{\small $\lambda$-ECLIPSE~\cite{patel2024lambda}} & 0.613 & 0.783 & 0.307 & 0.568 \\ 
{\small ELITE~\cite{wei2023elite}} & 0.652 & 0.762 & 0.255 & 0.556 \\
{\small IP-Adapter~\cite{ye2023ip}} & 0.608 & 0.809 & 0.274 & 0.564 \\
{\small IP-Adapter-Plus~\cite{ye2023ip}} & 0.692 & 0.826 & 0.281 & 0.600 \\
{\small SSR-Encoder~\cite{zhang2024ssr}} & 0.612 & 0.821 & 0.308 & 0.580 \\
{\small BLIP-Diffusion~\cite{li2023blip_diffusion}} & 0.594 & 0.779 & 0.300 & 0.558 \\
{\small MS-Diffusion~\cite{wang2024ms}} & 0.671 & 0.792 & \textbf{0.321} & 0.595 \\
{\small Subject-Diffusion~\cite{ma2024subject_diffusion}} & 0.711 & 0.787 & 0.293 & 0.597 \\
{\small JeDi~\cite{zeng2024jedi}} & 0.679 & 0.814 & 0.293 & 0.595 \\

\midrule
{\small \bf PatchDPO} & \textbf{0.727} & \textbf{0.838} & 0.292 & \textbf{0.619} \\

\bottomrule
\end{tabular}}
\caption{Performance comparison for single-object personalized generation on \textit{DreamBench}. The upper methods are finetuning-based methods, the bottom methods are finetuning-free methods, and bold font denotes the best result. ``SD'' is Stable Diffusion.}
\vspace{-1em}
\label{tab:benchmark_dream_single}

\end{table}

\vspace{0.5em}
\noindent \textbf{Implementation details.}
Our main experiments are conducted on the pre-trained IP-Adapter-Plus~\cite{ye2023ip} with SDXL model~\cite{podell2023sdxl} as the text-to-image diffusion model and OpenCLIP ViT-H/14 as the image encoder.
Note that IP-Adapter-Plus is the advanced version of the original IP-Adapter with significantly superior performance, by using the Resampler~\cite{jaegle2021perceiver} to extract reference image features.
This work only estimates the patch quality of object in the image to eliminate the interference from the background.
The parameters of the SDXL model and image encoder are frozen, and only the parameters for projecting image features are trainable.
During training, we adopt AdamW optimizer with a learning rate of 3e-5, and train the model on 8 GPUs for 30,000 steps with a batch size of 4 per GPU.
Besides, the self-supervised training of patch feature extraction is conducted for 10 epochs with a learning rate of 1e-1.


\vspace{0.5em}
\noindent \textbf{Training dataset.}
This work constructs the training dataset as illustrated in \autoref{sec:data_construction}.
Detailedly, the datasets for single-object \& multi-object personalized generation both consist of 50,000 images.
More details of multi-object personalized generation are in {\bf\color{red} S2.1} of the appendix.

\vspace{0.5em}
\noindent \textbf{Test benchmark.}
For single-object personalized image generation, we adopt the famous \textit{DreamBench}~\cite{ruiz2023dreambooth} as the benchmark.
For multi-object personalized image generation, we follow the \textit{Concept101}~\cite{kumari2023custom_diffusion} benchmark that has evaluated many methods.
Besides, \textit{MultiDreamBench}~\cite{ma2024subject_diffusion} is also adopted for comparison with Subject-Diffusion.

\vspace{0.5em}
\noindent \textbf{Evaluation metrics.}
We follow previous methods to adopt three metrics~(CLIP-T, CLIP-I, and DINO) for evaluation.
Specifically, CLIP-T evaluates the similarity between the generated images and given text prompts;
CLIP-I and DINO evaluate the similarity between the generated images and the reference images.
5 images are generated for each prompt to ensure the evaluation stability.
Besides, Avg.~(the average of three metrics) is also calculated for a comprehensive comparison.

\vspace{0.5em}
\noindent \textbf{Baseline methods.}
We compare our method with both finetuning-based methods~(\textit{e.g.}, Textual Inversion~\cite{gal2023textual_inversion}, DreamBooth~\cite{ruiz2023dreambooth}, Custom Diffusion~\cite{kumari2023custom_diffusion}) and finetuning-free methods~(\textit{e.g.}, SSR-Encoder~\cite{zhang2024ssr}, Subject-Diffusion~\cite{ma2024subject_diffusion}, JeDi~\cite{zeng2024jedi}).

\begin{table}
\renewcommand\arraystretch{1}
\small
\centering
\setlength{\tabcolsep}{0.80mm}{
\begin{tabular}{c *4{c}}
  \toprule
\textbf{\small Method} & \textbf{\small DINO} & \textbf{\small CLIP-I} & \textbf{\small CLIP-T}  & \textbf{\small Avg.} \\

\midrule

{\small Kosmos-G~\cite{pan2024kosmosg}}~(single image) & 0.694 & 0.847 & 0.287 & 0.609 \\
{\small Emu2-Gen~\cite{sun2024emu2}}~(single image) & 0.766 & 0.850 & 0.287 & 0.634 \\
{\small OmniGen~\cite{xiao2024omnigen}}~(single image) & 0.801 & 0.847 & \textbf{0.301} & 0.650 \\

\midrule
{\small \bf PatchDPO}~(single image) & \textbf{0.831} & \textbf{0.880} & 0.288 & \textbf{0.666} \\

\bottomrule
\end{tabular}}
\caption{Performance comparison for single-object personalized generation on \textit{DreamBench} using the evaluation setting of Kosmos-G. In this setting, only one image is preserved for each object in \textit{DreamBench}.}
\vspace{-1em}
\label{tab:benchmark_dream_single_kosmosg}

\end{table}
\begin{figure*}[t]
\centering
    \includegraphics[width=\linewidth]{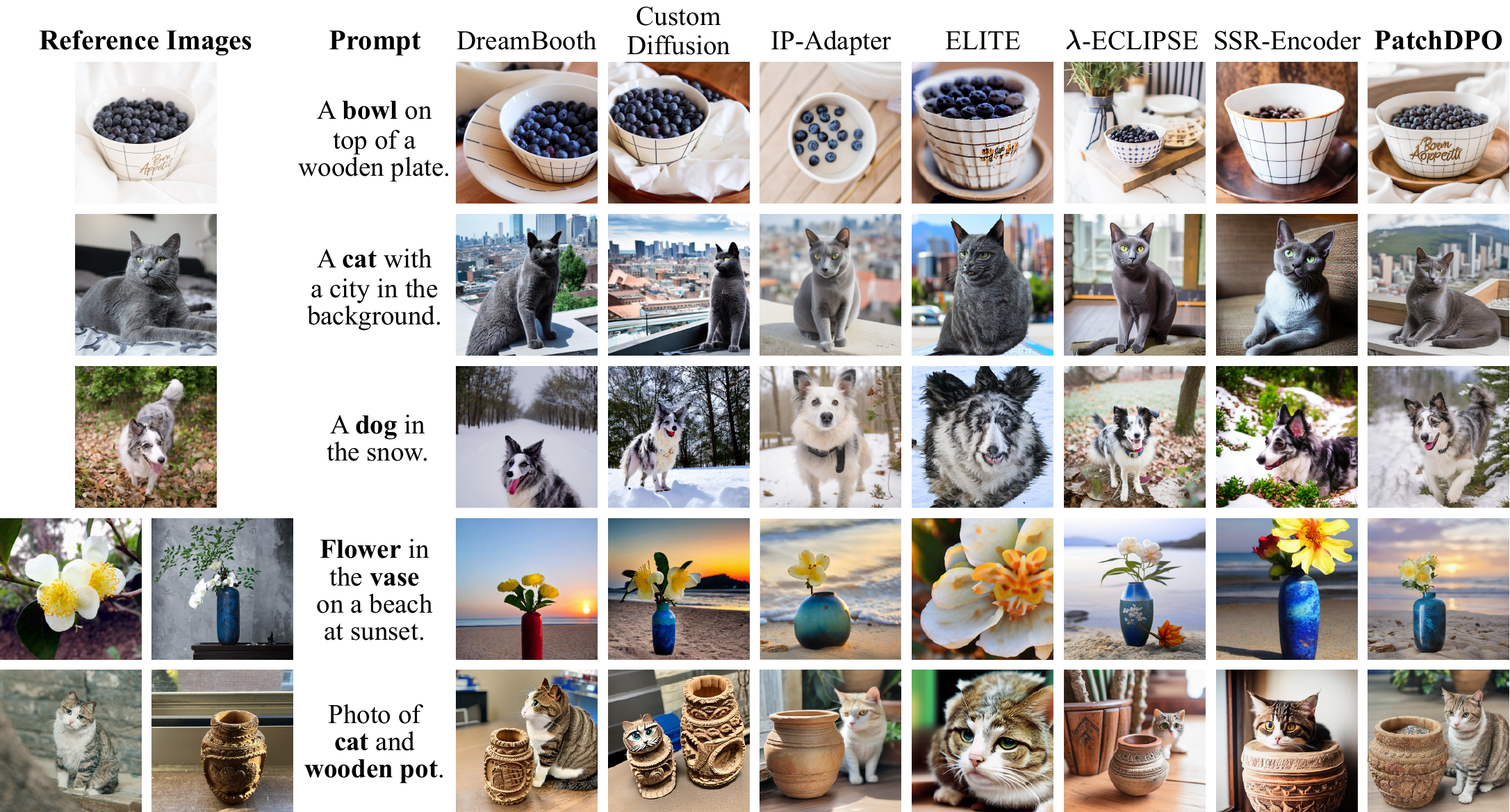}
\caption{
Qualitative comparisons of different methods on single-object \& multi-object personalized image generation.
}
\vspace{-1em}
\label{fig:compared_imgs}
\end{figure*}

\subsection{Single-Object Personalized Generation}

We conduct both quantitative and qualitative comparisons between our method and baseline methods.

\vspace{0.5em}
\noindent \textbf{Quantitative comparisons.}
\autoref{tab:benchmark_dream_single} \& \autoref{tab:benchmark_dream_single_kosmosg} demonstrates the quantitative results of different methods on \textit{DreamBench}.
Note that \autoref{tab:benchmark_dream_single} uses the original setting following most existing methods, where DINO, CLIP-I are calculated by comparing the generated image and \textbf{all images of the same object}.
\autoref{tab:benchmark_dream_single_kosmosg} uses the evaluation setting following Kosmos-G~\cite{pan2024kosmosg}, where only one image is preserved for each object, and DINO, CLIP-I are calculated by comparing the generated image and \textbf{only this image}.
The results of baseline methods are from their paper.

As shown in \autoref{tab:benchmark_dream_single} \& \autoref{tab:benchmark_dream_single_kosmosg}, PatchDPO realizes significantly superior image similarity~(DINO, CLIP-I) to the SOTA personalized generation methods, because PatchDPO provides very detailed patch-level feedback on the model's generated images, and facilitates the model to correct the low-quality patches that are inconsistent with those from the reference images.
Furthermore, PatchDPO achieves text similarity~(CLIP-T) comparable to SOTA methods, because each pair of reference images and generated images in the training dataset is from the same text prompt.
Therefore, aligning the low-quality patches of generated images with reference images does not decrease the similarity between the generated images and the text prompt.
Finally, our method also surpasses existing methods in average performance~(\textbf{Avg.}) by a large margin.

\vspace{0.5em}
\noindent \textbf{Qualitative comparisons.}
The upper part of \autoref{fig:compared_imgs} demonstrates the qualitative results of different methods on \textit{DreamBench}.
Compared to existing methods, PatchDPO excels in preserving the local details of the reference image, thus achieving generation of higher quality.




\vspace{0.5em}
\noindent \textbf{Qualitative comparisons.}
The bottom of \autoref{fig:compared_imgs} presents the qualitative results of different methods on \textit{Concept101}, showing that PatchDPO can better preserve the details of reference images in multi-object personalized generation.

Furthermore, the experiment results on \textit{Concept101} and \textit{MultiDreamBench}~(see {\bf\color{red} S2.2} of the appendix) show that PatchDPO can also improve the performance of original IP-Adapter-Plus on multi-object personalized generation, and achieves superior performance to existing personalized generation methods on both \textit{Concept101} \& \textit{MultiDreamBench}.

\subsection{Ablation Experiments}

We conduct the main ablation experiments of PatchDPO on \textit{DreamBench}, as demonstrated in \autoref{tab:ablation_experiments}.

\vspace{0.5em}
\noindent \textbf{Training datasets.}
This work compares our training dataset $\mathcal{D}_{\rm ours}$ of 50,000 images constructed in \autoref{sec:data_construction} with a natural dataset $\mathcal{D}_{\rm natural}$.
$\mathcal{D}_{\rm natural}$ consists of also 50,000 images~(with one main object in the image) randomly selected from the open-source SA-1B dataset~\cite{kirillov2023sam}.

In \autoref{tab:ablation_experiments}, Combination (1) seriously degrades the image-alignment~(DINO, CLIP-I) of model and Combination (2) benefits the model performance, indicating that the low-quality natural images~(with chaotic object details \& confused foreground and background, as shown in {\bf\color{red} S3} of the appendix) hinder the training of personalized generation.
Note that $\mathcal{L}_{\rm mse}$ is the loss of original diffusion model.

\vspace{0.5em}
\noindent \textbf{Training strategies.}
This work compares three training strategies corresponding to three losses.
$\mathcal{L}_{\rm mse}$ is the loss of original diffusion model.
$\mathcal{L}_{\rm DPO}$ is the loss of traditional DPO. Detailedly, we leverage the Diffusion-DPO loss~\cite{wallace2024diffusion_dpo} that directly adapts the original DPO loss to diffusion model.
Finally, $\mathcal{L}_{\rm PatchDPO}$ is the loss of PatchDPO.

In \autoref{tab:ablation_experiments}, Combination (3)~(traditional DPO) fails to improve the performance of the original model, because traditional DPO would wrongly reward the low-quality patches in the superior sample, while wrongly punishing the high-quality patches in the inferior sample.
Instead, Combination (4)~(PatchDPO) correctly rewards the high-quality patches and punishes the low-quality patches, thus achieving a huge performance improvement.

\vspace{0.5em}
\noindent \textbf{Patch features extraction.}
This work estimates the extracted patch features with $S_{\rm patch}$ in \autoref{sec:patch_quality_estimation}, and here we compare the extracted patch features of low $S_{\rm patch}$~(68.4\%, from the last feature map of original vision model $f$) and high $S_{\rm patch}$~(83.7\%, from the shallow feature map of vision model $f$ after self-supervised training).

In \autoref{tab:ablation_experiments}, Combination (5)~(PatchDPO with high $S_{\rm patch}$) surpasses Combination (4)~(PatchDPO with low  $S_{\rm patch}$), implying that patch features with higher $S_{\rm patch}$ contribute to more precise patch quality estimation and provide more accurate feedback for the generation model.

\vspace{0.5em}
\noindent \textbf{Additional ablation experiments.}
Besides, we provide more ablation experiments~(\eg, PatchDPO on different personalized generation models) in {\bf\color{red} S2.3} of the appendix.

\begin{table}
\renewcommand\arraystretch{1}
\small
\centering
\setlength{\tabcolsep}{1.0mm}{
\begin{tabular}{c *3{c}}
  \toprule
\textbf{\small Combination} & \textbf{\small DINO} & \textbf{\small CLIP-I} & \textbf{\small CLIP-T} \\
\midrule

Original model & 0.692 & 0.826 & 0.281 \\ 
(1) $\mathcal{L}_{\rm mse}$ + $\mathcal{D}_{\rm natural}$ & 0.658 & 0.818 & 0.287 \\
(2) $\mathcal{L}_{\rm mse}$ + $\mathcal{D}_{\rm ours}$ & 0.708 & 0.830 & 0.292 \\
(3) $\mathcal{L}_{\rm DPO}$ + $\mathcal{D}_{\rm ours}$ & 0.676 & 0.819 & 0.284 \\
(4) $\mathcal{L}_{\rm PatchDPO}$ + $\mathcal{D}_{\rm ours}$ + Original $f$ & 0.719 & 0.835 & 0.291 \\
(5) $\mathcal{L}_{\rm PatchDPO}$ + $\mathcal{D}_{\rm ours}$ + Trained $f$ & \textbf{0.727} & \textbf{0.838} & \textbf{0.292} \\ 

\bottomrule
\end{tabular}}
\caption{Ablation experiments.}
\vspace{-1em}
\label{tab:ablation_experiments}

\end{table}
\begin{figure}[t]
\centering
    \includegraphics[width=\linewidth]{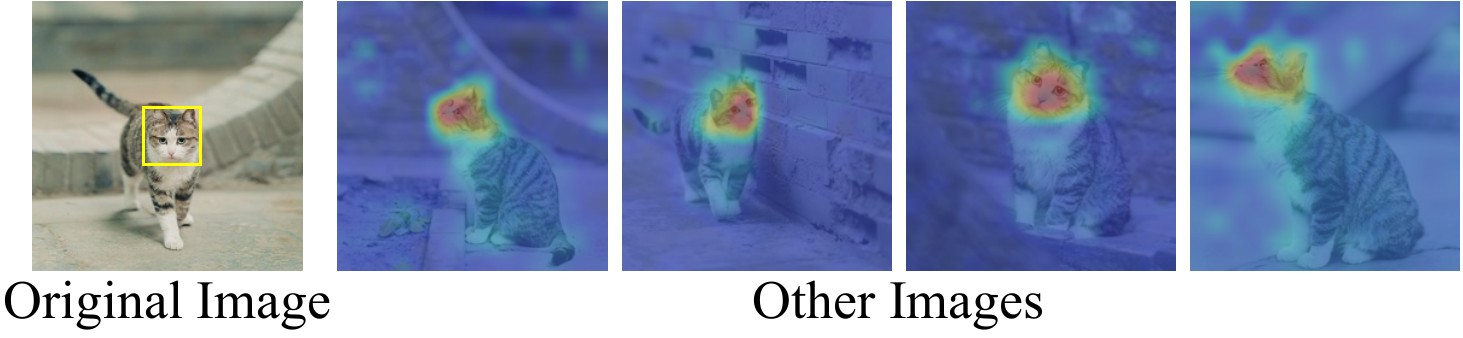}
\caption{
The matching heatmaps of target patch~(cat's head in the original image) on other images of the same cat.
}
\vspace{-1em}
\label{fig:matching_heatmaps}
\end{figure}

\subsection{Visualization Analysis}

\vspace{0.5em}
\noindent \textbf{Patch matching of extracted patch features.}
Here, this work visualizes the patch matching results of the extracted patch features.
Specifically, for the target patch in the original image, we acquire a matching heatmap $z \in \mathbb{R}^{H \times W}$ by calculating the similarity between its features and all patch features of another image.
Note that $z[h, w] \in \mathbb{R}$ indicates the similarity between the target patch and the patch in the $h$-th row and the $w$-th column of another image.
Next, we visualize $z$ by resizing it to the same size as the image, and overlapping them.
As shown in \autoref{fig:matching_heatmaps}, the matching heatmap $z$ accurately highlights the correct patch corresponding to the target patch, implying that the extracted patch features accurately represent the corresponding patch.

\begin{figure}[t]
\centering
    \includegraphics[width=\linewidth]{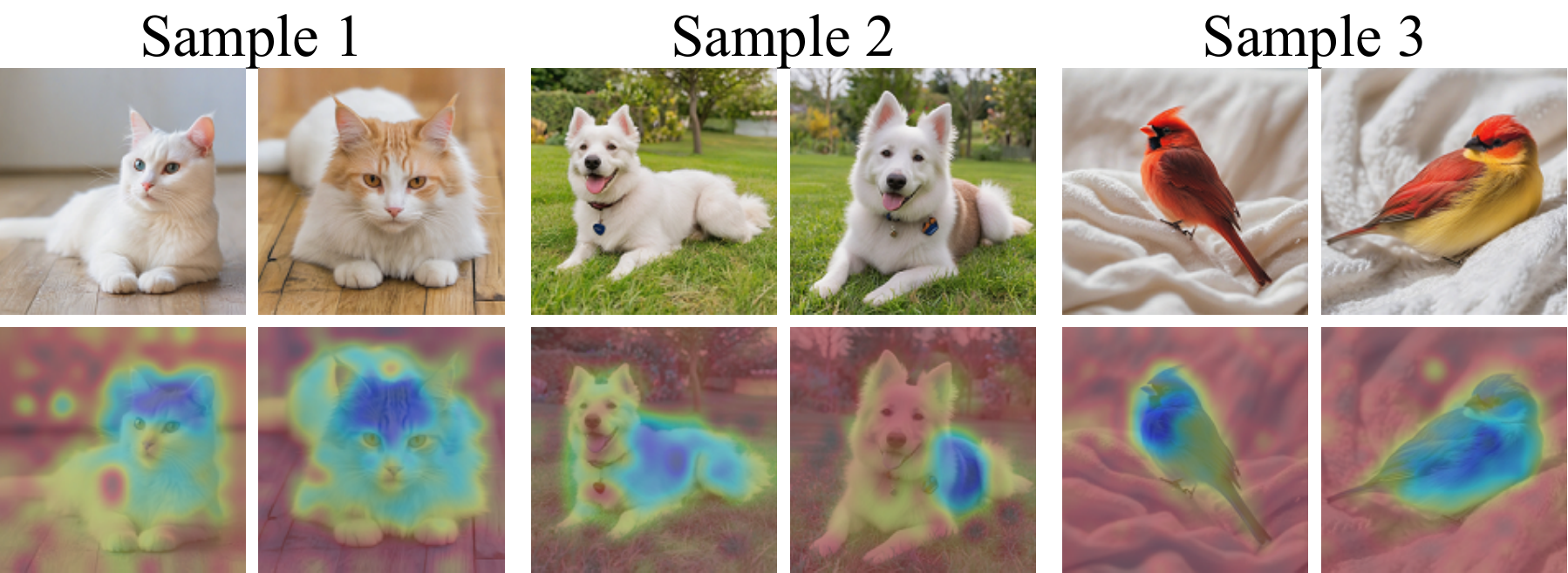}
\caption{
Three samples of patch quality estimation.
}
\vspace{-1em}
\label{fig:patch_quality}
\end{figure}

\vspace{0.5em}
\noindent \textbf{Patch quality estimation.}
Here, we visualize the estimated patch quality, $p(\mybold{x}_{\rm ref}) \in \mathbb{R}^{H \times W}$ and $p(\mybold{x}_{\rm gen}) \in \mathbb{R}^{H \times W}$, in the same manner as visualizing the matching heatmap $z$.
As demonstrated in \autoref{fig:patch_quality}, the dark patches in the image are inconsistent with the corresponding patches in another image, indicating that our method can accurately estimate the patch quality.

\begin{figure}[t]
\centering
    \includegraphics[width=\linewidth]{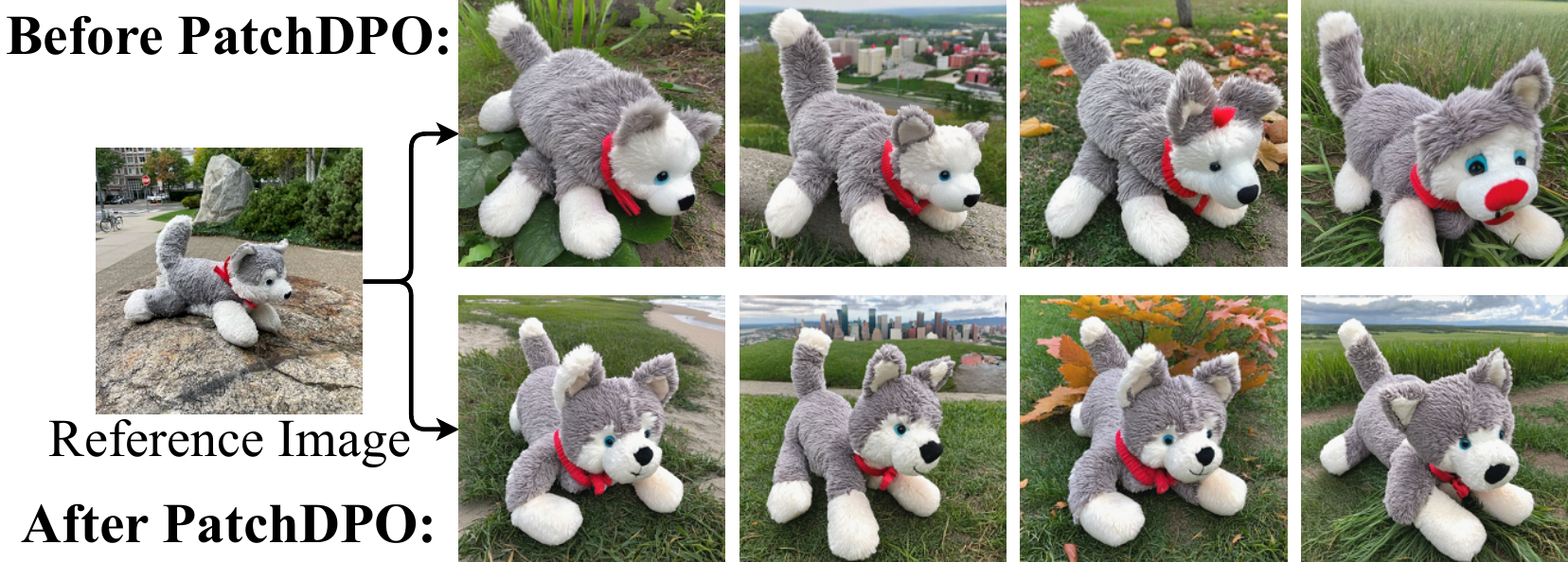}
\caption{
Qualitative ablation experiment.
}
\vspace{-1em}
\label{fig:ablation_imgs}
\end{figure}

\vspace{0.5em}
\noindent \textbf{Images before/after PatchDPO.}
Here, this work demonstrates the generated images from the model before/after the PatchDPO training.
As shown in \autoref{fig:ablation_imgs}, the images from the model after PatchDPO exhibit significantly enhanced quality in terms of image-alignment, highlighting the effectiveness of PatchDPO.

\vspace{0.5em}
\noindent \textbf{Additional visualization results.}
Furthermore, we provide more visualization results in {\bf\color{red} S3} of the appendix for a comprehensive understanding of our method.

\section{Conclusion}

In this work, we propose PatchDPO~(patch-level direct preference optimization), which leverages an additional training stage to improve the pre-trained personalized generation models.
PatchDPO estimates the quality of image patches within each generated image and accordingly provides detailed feedback to the models.
Specifically, we propose a patch quality estimation method based on the pre-trained vision model finetuned with a self-supervised training method.
Next, we propose a weighted training approach to train the model with the estimated patch quality, which rewards high-quality image patches while penalizing those of low quality.
Experiment results demonstrate that PatchDPO achieves state-of-the-art performance on both single-object and multi-object personalized image generation.
We hope our method and dataset can contribute to the community of personalized image generation.

\vspace{0.5em}
\noindent \textbf{Acknowledgements.}
This work is supported by Zhejiang Province High-Level Talents Special Support Program ``Leading Talent of Technological Innovation of Ten-Thousands Talents Program''~(No. 2022R52046), and Alibaba-Zhejiang University Joint Research Institute of Frontier Technologies.

{
    \small
    \bibliographystyle{ieeenat_fullname}
    \bibliography{main}
}


\end{document}